\documentclass[times,twocolumn,final]{elsarticle}

\usepackage{medima}
\usepackage{framed,multirow}

\usepackage{amssymb}
\usepackage{latexsym}

\usepackage{url}
\usepackage{xcolor}

\usepackage{epsfig}
\usepackage{graphicx}

\usepackage{bbding} 
\usepackage{pifont} 
\usepackage{wrapfig} 
\usepackage{microtype}
\usepackage{amsfonts}
\usepackage{soul}

\usepackage{amsmath,amssymb} 
\usepackage{mathrsfs} 
\usepackage{algorithmic,algorithm}
\usepackage{url} 
\usepackage{multirow} 
\usepackage{subfigure} 
\usepackage{bbding} 
\usepackage{pifont} 
\usepackage{wrapfig}  
\usepackage{microtype} 
\usepackage{amsfonts} 
\usepackage{array,caption}
\usepackage{bbm}
\usepackage{xcolor,colortbl}
\usepackage{soul}
\usepackage{microtype}
\usepackage{booktabs}
\usepackage{multirow}
\usepackage{makecell} %
\usepackage{tikz}

\usepackage{url}  
\usepackage[colorlinks, linkcolor=black, anchorcolor=black, citecolor=black]{hyperref}

\newcommand{\cmark}{\ding{51}}%
\newcommand{\xmark}{\ding{55}}%

\newcommand{\changed}[1]{\textcolor{black}{#1}}

\newcommand*\circled[1]{\tikz[baseline=(char.base)]{
            \node[shape=circle,draw,inner sep=1pt] (char) {#1};}}

\def\shortmodelname{DSPNet}
\def\shortnamefspa{FSPA}
\def\shortnameblna{BCMA}

\definecolor{cmblu}{RGB}{51,102,240}
\definecolor{cmred}{RGB}{241,22,22}

\definecolor{newcolor}{rgb}{.8,.349,.1}

\journal{Medical Image Analysis}

\begin{document}

\verso{Song Tang \textit{et~al.}}

\begin{frontmatter}

\title{Few-Shot Medical Image Segmentation with High-Fidelity Prototypes}

\author[1,2]{Song \snm{Tang}}
\cortext[cor1]{Corresponding author: Mao Ye~(cvlab.uestc@gmail.com) and Xiatian Zhu~(xiatian.zhu@surrey.ac.uk).}
\author[1]{Shaxu   \snm{Yan}}
\author[5]{Xiaozhi \snm{Qi}}
\author[1]{Jianxin \snm{Gao}}
\author[3]{Mao     \snm{Ye}\corref{cor1}}
\author[2]{Jianwei \snm{Zhang}}
\author[4]{Xiatian \snm{Zhu}\corref{cor1}}

\address[1]{IMI Group, School of Health Sciences and Engineering, University of Shanghai for Science and Technology, Shanghai, China} 
\address[2]{TAMS Group, Department of Informatics, Universität Hamburg, Hamburg, Germany}
\address[3]{School of Computer Science and Engineering, University of Electronic Science and Technology of China, Chengdu, China}
\address[4]{Surrey Institute for People-Centred Artificial Intelligence, and Centre for Vision, Speech and Signal Processing, University of Surrey, Guildford, UK}
\address[5]{Shenzhen Key Laboratory of Minimally Invasive Surgical Robotics and System, Shenzhen Institute of Advanced Technology, Chinese Academy of Sciences, China}

\received{*}
\finalform{*}
\accepted{*}
\availableonline{*}

\begin{abstract}
Few-shot Semantic Segmentation (FSS) aims to adapt a pretrained model to new classes with as few as a single labelled training sample per class. Despite the prototype based approaches have achieved substantial success, existing models are limited to the imaging scenarios with considerably distinct objects and not highly complex background, e.g., natural images. This makes such models suboptimal for medical imaging with both conditions invalid. To address this problem, we propose a novel {\it {\bf D}etail {\bf S}elf-refined {\bf P}rototype {\bf Net}work} ({\bf {\shortmodelname}}) to constructing high-fidelity prototypes representing the object foreground and the background more comprehensively. Specifically, to construct global semantics while maintaining the captured detail semantics, we learn the foreground prototypes by modelling the multi-modal structures with clustering and then fusing each in a channel-wise manner. Considering that the background often has no apparent semantic relation in the spatial dimensions, we integrate channel-specific structural information under sparse channel-aware regulation. Extensive experiments on three challenging medical image benchmarks show the superiority of DSPNet over previous state-of-the-art methods. 
The code and data are available at \url{https://github.com/tntek/DSPNet}. 
\end{abstract}

\begin{keyword}
\MSC *\sep *\sep *\sep *
\KWD Few-shot semantic segmentation \sep Medical image \sep High-fidelity prototype \sep Detail self-refining
\end{keyword}

\end{frontmatter}


\section{Introduction} \label{sec:introduction}
Medical image segmentation plays a critical role in clinical processes and medical research, such as disease diagnosis~\citep{zhu2022multimodal}, treatment planning~\citep{sherer2021metrics} and follow-up~\citep{r3}. 
In the medical field, the well-annotated samples are limited due to privacy protection and the requirement of clinical expertise.
Within this context, Few-shot Semantic Segmentation~(FSS) methods~\citep{ouyang2022self} demonstrate their advantages in this domain, involving extracting one or several supporting data to predict the same type in query data. 
\begin{figure}[t] 
    \begin{center}
     \includegraphics[width=0.98\linewidth]{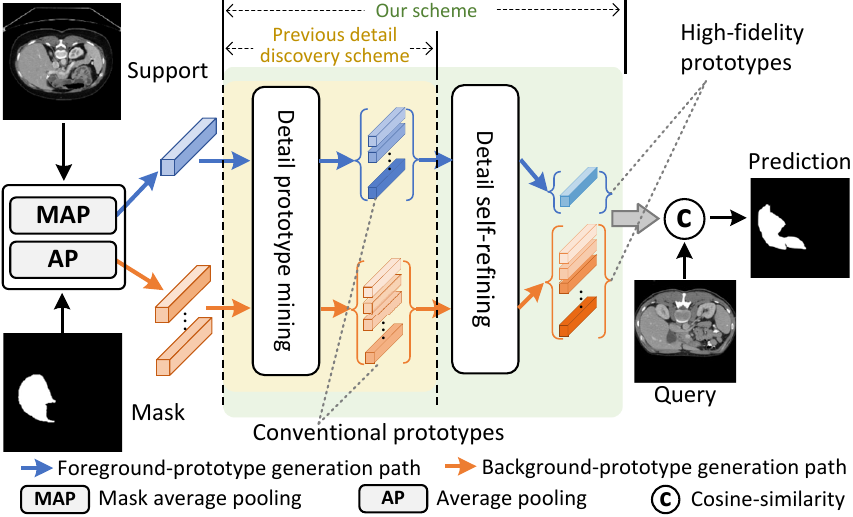}
    \end{center}
    \caption{
    Scheme comparison. 
    For the local information loss problem caused by pooling operation, {\it previous detail discovery scheme} incrementally mines new prototypes to capture more details.
    Our scheme is featured with the design called {\it detail self-refining}, aiming at encouraging high-fidelity prototypes.
    To enhance the deep representation for details, our self-refining works in different way:   
    The foreground class prototype is refreshed by fusing the cluster-mined semantics, whilst background prototypes are enhanced by incorporating the channel-specific structural information. 
    }
    \label{fig:idea-comp}
\end{figure}


The key to FSS is building a resemblance between the support and query images. 
The existing FSS methods follow three clues. 
The first is constructing support images-based guidance to boost this query image segmentation, e.g., the two-branch architecture with interaction~\citep{shaban2017one, roy2020squeeze}.   
The second identifies the shared features by building a resemblance between the support and query images, e.g., attention modules~\citep{hu2019attention} and graph networks~\citep{gao2022mutually}.
The third is prototypical approaches~\citep{snell2017prototypical, ouyang2020self}, mining prototypes from support images to build a resemblance with the query images.
Among them, the third one is the current prevalent scheme due to the generality and robustness to noise.
However, given that the prototype extraction utilizes the pooling operation, e.g., Mask Average Pooling or Average Pooling, this scheme suffers from an inherent limitation: 
{\it Since the pooling is prone to losing local details, the conventional prototypes lead to low-discriminative feature maps that confuse the foreground and background.}  



Existing methods address this above-mentioned limitation by incrementally mining new prototypes for diverse detail representations, i.e., the detail discovery scheme marked by a yellow zone in Fig.~\ref{fig:idea-comp}.
For instance, the single class prototype for foreground was enriched by several part-aware prototypes~\citep{liu2020part} or compensation prototypes~\citep{zhang2021self}. 
For background,  Average Pooling was employed at the regular grid to generate diverse local prototypes~\citep{ouyang2020self}. 
This strategy works well in imaging scenarios with (i) considerably distinct objects and (ii) not highly complex background, e.g., natural images. 
However, the medical images with highly heterogeneous textures\footnote{Refer to these considerable and complicated distinct structures/tissues with compact boundaries between them.} do not satisfy the conditions. 
Namely, the incremental strategy cannot provide complete detail representations for medical images.





To overcome this problem above, in this paper, we propose a new {{\bf D}etail {\bf S}elf-refined {\bf P}rototype {\bf Net}work} ({\bf {\shortmodelname}}).
As demonstrated in Fig.~\ref{fig:idea-comp} (see the green zone), in contrast to constructing new prototypes, our scheme highlights enhancing details representation of off-the-shelf prototypes by {\it detail self-refining}, leading to high-fidelity prototypes.  

In the proposed network, our detail self-refining involves two novel {\it attention-like} modules, called {Foreground Semantic Prototype Attention}~({\shortnamefspa}) and {Background Channel-structural Multi-head Attention}~({\shortnameblna}). 
In {\shortnamefspa}, to account for the clear semantics of foreground, we mine the semantic prototypes at the class level as the detail prototypes, using superpixel clustering. 
Then, they are fused to a single class prototype in a channel-wise one-dimensional convolution fashion, assembling the global semantics while maintaining the local semantics. 
In {\shortnameblna}, since the complicated background in medical images is often unsemantic, we take the ones generated by Average Pooling at the regular grid as the background detail prototypes, instead of mining detail information from the spatial dimension. 
Then, the channel-specific structural information is explored by combining learnable global information with an adjustment highlighting sparse-relative channels.
In the end, the elements of each detail prototype are channel-wise refreshed independently by the corresponding channel-specific structural information.

The \textbf{contributions} of this work are three folds. 
We propose: 


\begin{itemize}
\item[(1)] A novel prototypical FSS approach {\shortmodelname} that enhances prototypes' self-representation for complicated details, totally discrepant from the previous incremental paradigm of constructing new detail prototypes.

\item[(2)] A self-refining method {\shortnamefspa} for class prototype that integrates the cluster prototypes, i.e., the mined semantic details, into an enhanced one in an attention-like fashion and indicates the potential of fusing cluster-based local details for complete foreground representation.  

\item[(3)] A self-refining method {\shortnameblna} for background prototype that \changed{incorporates the channel-specific structural information by multi-head channel attention with sparse channel-aware regularization} and provides a conceptually different view for background details modeling.  
\end{itemize}

\section{Related Work}\label{sec:rework}
\subsection{Medical Image Segmentation}
Currently, the deep neural network approaches dominate the medical image segmentation field.
The early phase shares models with the natural image semantic segmentation. 
Fully Convolutional Networks (FCNs)~\citep{long2015fully} first equipped vanilla Convolutional Networks (CNN) with a segmentation head by introducing Up-sampling and Skip layer. 
Following that, the encoder-decoder-based methods~\citep{badrinarayanan2017segnet, noh2015learning} are developed. 
Unlike the coarse reconstruction in FCNs, the symmetrical reconstruction of the decoder can capture much richer detailed semantics. 
With the application of deep learning in the medical field, the medical image-specific models merge correspondingly, among which U-Net~\citep{noh2015learning} is extensively recognized for its superior performance.  
Besides symmetrical encoder-decoder architecture, U-Net infuses the skipped connections to facilitate the propagation of contextual information to higher resolution hierarchies.
Inspired by it, several variants of U-Net are designed, including U-Net 3D~\citep{cciccek20163d}, Atten-U-Net~\citep{oktay2018attention}, Edge-U-Net~\citep{allah2023edge}, V-Net~\citep{milletari2016v} and Y-Net~\citep{mehta2018net}.

These segmentation models mentioned above only work in a supervised fashion, relying on abundant expert-annotated data. Thus, they cannot apply to the few-shot setting where we need to segment an object of an "unseen" class as only a few labeled images of this class are given.

\subsection{Few-shot Semantic Segmentation~(FSS)} 
The key to solving FSS is building a class-wise similarity between the query and support images. 
Following this view, the existing methods can be divided into three categories. 
The first category constructs support images-based guidance~\citep{shaban2017one,roy2020squeeze,wu2022dual,sun2022few}. 
For instance, \citep{shaban2017one} developed a two-branch approach where a conditioning branch imposes controlling on the logistic regression layer of the segmentation branch. 
\citep{roy2020squeeze} introduced squeeze and excitation blocks into the conditioning branch to encourage dense information interaction between the two branches. 
The second category designed novel network modules, e.g., attention modules~\citep{hu2019attention,wang2020few} and graph networks~\citep{gao2022mutually,xie2021scale}, for discriminative representations, by which the features shared by query and support images were identified. 
The third mainstream is prototypic network~\citep{wang2019panet,ouyang2020self,ding2023self, feng2021interactive} that prototypes bridge the similarity computation in a meta-learning fashion. 
Here, the prototypes are specific features with semantics extracted from the support images. 
Recently, PANet~\citep{wang2019panet} achieved impressive performance on the natural image segmentation task, performing dual-directive alignment between the query and support images. 
SSL-PANet~\citep{ouyang2020self} transferred the PANet architecture into the medical image segmentation where self-supervision with superpixels and local representation ensure the unsupervised segmentation. 
Following that, anomaly detection-inspired methods enhanced the performance by introducing self-supervision with supervoxels\citep{hansen2022anomaly} or learning mechanism of an expert clinician~\citep{shen2023qnet}.

Our {\shortmodelname} belongs to the third category above, having two significant discrepancies. 
Compared with methods working with classes-abundant natural images, e.g., PANet, {\shortmodelname} is a medical image-specific model with limited labelled support data. 
On the other hand, {\shortmodelname} considers the limitation of local information loss from a new view of detail self-refining, which is totally different from the existing prototypical methods.

\subsection{Attention Method in Few-Shot Semantic Segmentation} 
For few-shot semantic segmentation tasks, attention mechanism~\citep{vaswani2017attention} is a popular technique to build the relationship between the support and query images. 
The existing approaches can be divided into two categories: (i) graph-based~\citep{zhang2019pyramid,wang2020few,gao2022mutually} and (ii) non-graph-based~\citep{hu2019attention,zhang2022catrans, mao2022task}.
The methods belonging to the first category employing a graph model to activate more pixels, such that the correspondence between support and query images is enhanced. 
For example, \citep{gao2022mutually} fused the graph attention and the last layer feature map to generate an enhanced feature map to solve the problem of foreground pixel loss in the attention map. 
The core idea of the second category methods is to build the correspondence based on feature interaction between the support and query data, e.g., multi-scale contextual features~\citep{hu2019attention}, affinity constraint~\citep{zhang2022catrans}, mix attention map~\citep{mao2022task}.

\changed{Unlike the spatial attention-based methods above, both {\shortnamefspa} and {\shortnameblna} in {\shortmodelname} are channel attention-like methods. For {\shortnamefspa}, the channel-wise fusion ensures the deeper semantic fusion from local to global, whilst in {\shortnameblna}, the detail self-refining relies on the channel structural information.  
}

\section{Problem Statement} \label{sec:fss} 

In the case of few-shot segmentation, the dataset includes two parts, the training subset $\mathcal{D}_{tr}$ (annotated by $\mathcal{Y}_{tr}$) and the test subset $\mathcal{D}_{te}$ (annotated by $\mathcal{Y}_{te}$), both of which consist of image-mask pairs.
Furthermore, the $\mathcal{D}_{tr}$ and $\mathcal{D}_{te}$ do not share categories. 
Namely, $\mathcal{Y}_{tr}\cap \mathcal{Y}_{te}= \varnothing$. 
The goal of few-shot semantic segmentation is to train a segmentation model on $\mathcal{D}_{tr}$ that can segment unseen semantic classes $\mathcal{Y}_{te}$ in images in $\mathcal{D}_{te}$, given
a few annotated examples of $\mathcal{Y}_{te}$, without re-training.

\begin{figure*}[t]
    \begin{center}
        \includegraphics[width=0.95\linewidth,angle=0]{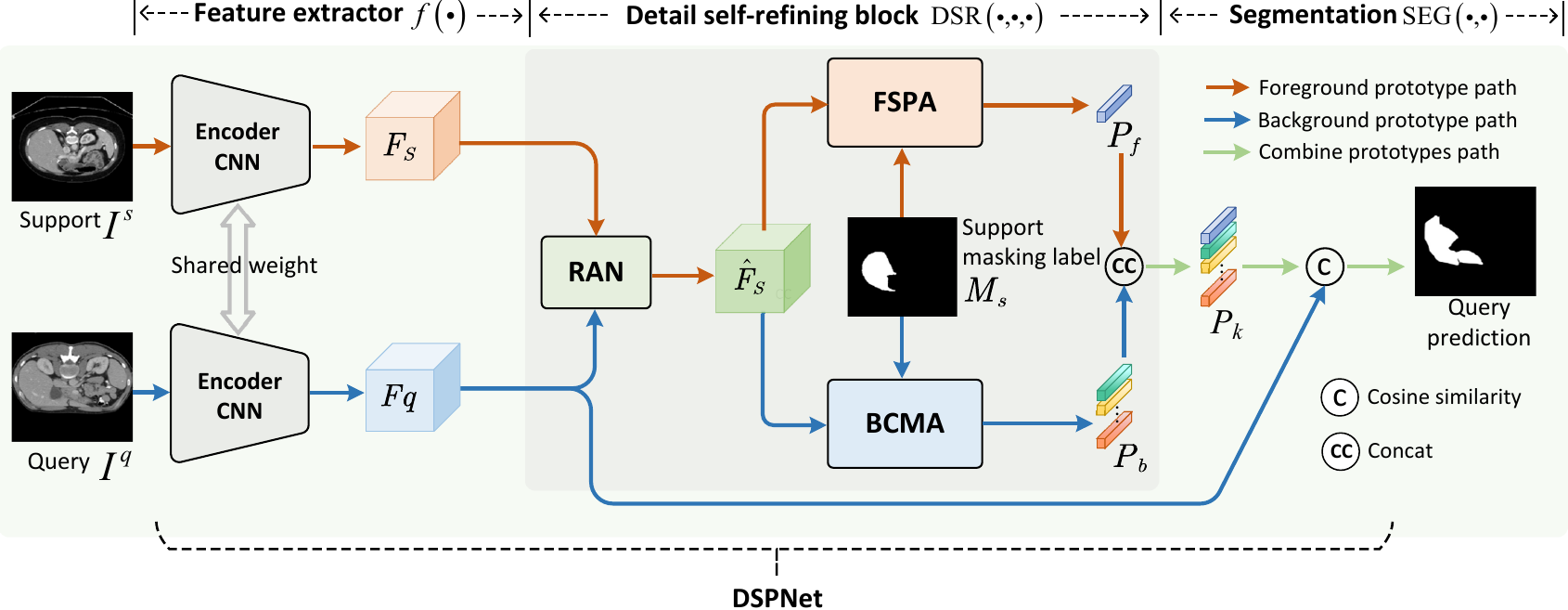}  
    \end{center} 
    \caption{
    Overview of the proposed {\shortmodelname}. 
    The segmentation pipeline of {\shortmodelname} follows three steps sequentially: 
    The feature extractor $f(\cdot)$ embeds the support image $I^{s}$ and query image $I^{q}$into deep features $F_{s}$ and $F_{q}$ respectively; after that, the prototypes are generated by the detail self-refining block $P_k={\rm{DSR}}(F_{s},F_{q},M_s)$; finally, the query image is segmented by measuring the cosine similarity between each prototype and query features at each pixel. 
    In the block of ${\rm{DSR}}(\cdot,\cdot,\cdot)$, RAN calibrates $F_{s}$, $F_{q}$ to filter irrelevant object and noise, and then, high-fidelity class prototype and background prototypes are generated via {\shortnamefspa} and {\shortnameblna}, respectively.  
    }
    \label{fig:fw}
\end{figure*}

To reach the goal above, we formulate this problem in a meta-learning fashion, the same as the initial few-shot semantic segmentation work. 
Specifically, 
$\mathcal{D}_{tr}= \{S_{i},Q_{i}\}_{i=1}^{N_{tr}}$ and $\mathcal{D}_{te}= \{S_{i},Q_{i}\}_{i=1}^{N_{te}}$ are sliced into several randomly sampled episodes, where ${N_{tr}}$ and ${N_{te}}$ are the episodes number for training and testing, respectively. 
Each episode consists of K annotated support images and a collection of query images containing N categories. 
Namely, we consider an N-way K-shot segmentation sub-problem. 
Specifically, the support set $S_{i}= \{ ({I}_{k}^{s},{m}_{k}^{s}({c}_{j})) \}_{k=1}^{K}$ contains K image-mask pairs of a gray-scale image $I\in {R}^{H\times W }$ and its corresponding binary mask ${m\in \{ {0,1} \}}^{H\times W}$ for class ${c}_{j}\in {C}_{tr},j=1,2,\cdots,N$.
The query set ${Q_{i}}$ contains ${V}$ image-mask pairs from the same class as the support set. 
While the training on $\mathcal{D}_{tr}$, over each episode, we learn a function $f(I^q, S_i)$, which predicts a binary mask of an unseen class when given the query image $I^q \in {Q_{i}}$ and the support set $S_i$.  
After a series of episodes, we obtain the final segmentation model, which is evaluated on ${N_{te}}$ in the same N-way K-shot segmentation manner. 
Following the common practice in~\citep{ouyang2020self, achanta2012slic, shen2023qnet}, this paper set $N=K=1$.


\section{Methodology} \label{sec:method}
In this work, we propose a detail representation enhanced network ({\shortmodelname}) for prototypical FSS, building on the self-supervision framework~\citep{ouyang2020self}.   
As shown in Fig.~\ref{fig:fw}(a), {\shortmodelname} consists of three modules from left to right: (i) The CNN-based feature extractor $f(\cdot)$; (ii) the detail self-refining block ${\rm{DSR}}(\cdot)$; and (iii) the segmentation block based on the cosine similarity. 
Suppose the support and query images are denoted by $I_s$ and $I_q$, respectively. 
The segmentation begins with feature extraction $F_s=f_{\theta}(I_s)$ and $F_q=f_{\theta}(I_q)$. 
Furthermore, high-fidelity foreground prototype and background prototypes are produced by the detail self-refining block, denoted by $P_k=\{P_f, P_b\}={\rm{DSR}}(F_q, F_s, M_s)$ where $M_s$ is the support masking label. 
Finally, we obtain the query prediction of segmentation ${\rm{SEG}}(F_q, P_k)$, computing cosine similarity between $F_q$ and obtained prototypes $P_k$ in a convolution fashion.

In the segmentation process above, the optimal prototype generation encouraged by the detail self-refining block, i.e., ${\rm{DSR}}(\cdot, \cdot, \cdot)$, distinguishes our {\shortmodelname} from the previous work.  
As shown in Fig.~\ref{fig:fw}(b), after RAN calibrates $F_s$ and $F_q$ to a semantics fused feature maps $\hat{F}_s$, {\shortnamefspa} and {\shortnameblna} extract cluster-based prototypes and Average Pooling-based prototypes from $\hat{F}_s$ respectively and take them as raw detail prototypes.  
Then, the high-fidelity class prototype $P_f$ and background prototypes $P_b$ are further obtained by the channel-wise fusion in {\shortnamefspa} and the sparse channel-aware multi-head channel attention in {\shortnameblna}. 
In the rest of this section, we will elaborate on the three key components.

\begin{figure}[t]
    \begin{center}
	\includegraphics[width=0.98\linewidth]{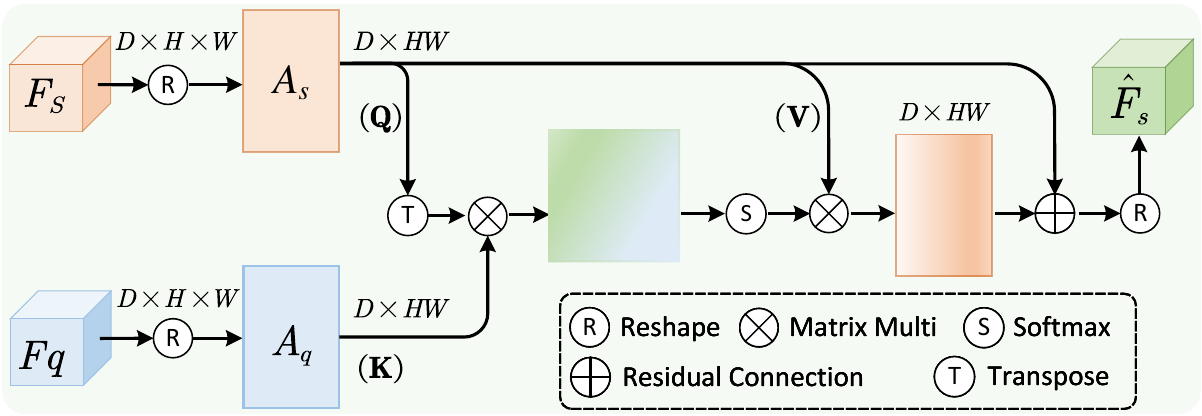}
    \end{center}
    \caption{Architecture of Resemblance Attention Network.} 
    \label{fig:ran}
\end{figure}

\subsection{Resemblance Attention Network} \label{sec:ran}
In the FSS field, Resemblance Attention Network (RAN) \citep{wang2018non} is a classic module to integrate the support and query features~\citep{fu2019dual,huang2019ccnet,ding2023few}.
In the proposed {\shortmodelname}, RAN engages in filtering irrelevant texture and objects between $F_s$ and $F_q$.   
Fig.~\ref{fig:ran} presents its network architecture. 
When support and query feature maps $F_s$, $F_q$ are input, they are first reshaped to feature vector $A_s$ and $A_q$, respectively. 
After that, in a {\bf Q}uery-{\bf K}ey-{\bf V}alue attention manner with residual connection, the $A_s$, $A_q$ are fused to $\hat{F}_s$ where ${\rm{\bf Q}}={\rm{\bf V}}=A_s$, ${\rm{\bf K}}=A_q$. The process can be formulated by Eq.~\eqref{eqn:ran}.
\begin{equation}\label{eqn:ran}
    \begin{split} 
    \hat{{F}}_{s} =  \frac{{{\phi }\left({A}_{s}^{T}\times {A}_{q}\right)}\times{A}_{s}}{\left \| {A}_{s}\right \|\left \|{A}_{q} \right \|}+ {A}_{s} .  
  \end{split}
\end{equation}
where $\phi(\cdot)$ stands for softmax operation, $\times$ means matrix multiplication, ${\phi }\left({A}_{s}^{T}\times {A}_{q}\right)$ means the similarity-based probability matrix weighting $A_s$.  




\subsection{Foreground Semantic Prototype Attention} \label{sec:fspa}

To obtain high-fidelity class prototype for the semantic foreground, we explore the local semantics in the foregroud and fuse them to form global semantics without local semantics loss. 
We accomplish this idea using the cluster-based detail prototypes and channel-wise attention with local semantics guidance.

\begin{figure}[t]
    \begin{center}
	\includegraphics[width=0.98\linewidth]{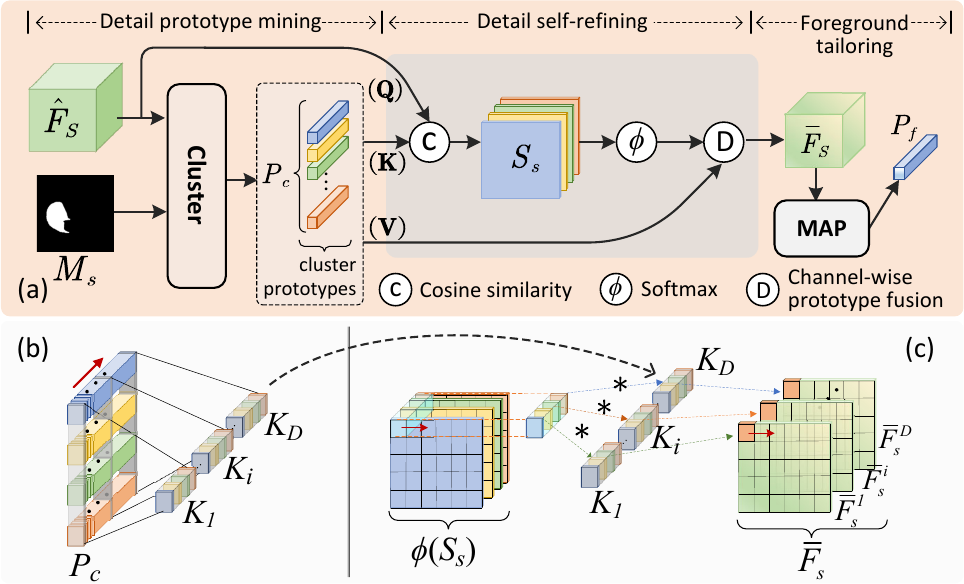} 
    \end{center}
    \caption{{\shortnamefspa} illustration. 
    {\bf (a)} shows {\shortnamefspa} architecture where both cosine similarity computation \textcircled{\tiny{C}} and channel-wise prototype fusion \textcircled{\tiny{D}} are implemented in a one-dimensional convolution manner. 
    For \textcircled{\tiny{C}}, the cluster prototypes $P_c$ serves as convolution filters individually. 
    Regarding \textcircled{\tiny{D}}, the channel-wise convolution fillers are generated from $P_c$ by channel-dimensional slicing (see (b)), whilst the prototype fusion over probability maps $\phi(S_s)$ is demonstrated in (c).} 
    \label{fig:fore}
\end{figure}

{\bf Overview}. 
As shown in Fig.~\ref{fig:fore}(a), to get more local semantics, we first employ the superpixel-guided clustering method~\citep{li2021adaptive} to mine $N_s$ cluster prototypes, denoted by ${P}_{c} =\{{P}_c^i\}_{i=1}^{N_s}$ where ${P}_{c}^{i} \in {\mathbb{R}}^{1 \times D}$, $D$ is the dimension of prototype. 
The intuitive fusion, e.g., vanilla weighting without prior knowledge~\citep{li2021adaptive}, can obtain the global semantics but suffers from confusing detail semantics. 
Therefore, we propose an \textit{attention-like cluster prototype fusion} to address this issue, implementing detail self-refining and foreground tailoring sequentially.

{\bf Attention-like cluster prototype fusion}. 
As shown in the middle of Fig.~\ref{fig:fore} (marked by grey box), this attention can be implemented in the fashion of {\bf Q}uery-{\bf K}ey-{\bf V}alue.
Taking ${\rm{\bf Q}}={\hat{F}_s}$, ${\rm{\bf K}}={\rm{\bf V}}={{P}_c}$, we can summarize this module to the following equation. 
\begin{equation}\label{eqn:fore-fusion} 
    \begin{split} 
    {\bar{F}}_s= {{\phi }\left(\hat{F}_s~{\scriptsize\circled{C}}~{P}_{c}\right)}~{\scriptsize\circled{D}}~{P}_{c}, 
    \end{split}
\end{equation} 
where $\phi(\cdot)$ is softmax computation; 
operator ${\scriptsize\circled{C}}$ and ${\scriptsize\circled{D}}$ respectively means the computation for cosine similarity measurement and channel-wise prototype fusion, whose details are presented in the following.

Since the size of $\hat{F}_s$ and $P_c$ are different, computation of ${\scriptsize\circled{C}}$ does not follow cosine similarity's definition, but performing in prototype-wise. 
Specifically, each prototype in ${P}_{c}$, i.e., ${P}_{c}^i$, is used to compute similarity with the supporting feature maps $\hat{F}_s$ in a one-dimensional convolution manner, in which the convolution calculation is replaced by cosine similarity computation. 
Thus, the $N_s$ prototypes lead to $N_s$ similarity maps, which can be collectively written as $S_s=\hat{\rm{F}}_s~{\scriptsize\circled{C}}~{P}_{c} \in {\mathbb{R}}^{(H_s \times W_s) \times N_s}$ where $(H_s, W_s)$ is map size. 
For any map in $S_s$, denoted by $S_s^i$, its computation can be expressed as   
\begin{equation}\label{eqn:cov1d-ss}
    \small
    \begin{split} 
    {{S}_s^i} = {\rm{sim1D}}~(\hat{F}_s, {P}_{c}^i),   
    \end{split}
\end{equation}
where function ${\rm{sim1D}}~(\cdot,\cdot)$ stands for the similarity computation working in the one-dimensional convolution fashion.   
The value of ${{S}_s^i}$ at position $(h,w)$ is the cosine similarity between ${P}_{c}^{i}$ and ${\hat{F}}_{s}$ at position $(h,w)$. Namely, 
\begin{equation}\label{eqn:cos-sim-xy}
    \begin{split} 
    {S}_s^{i}(h,w)=\frac{\left({P}_{c}^{i}\right)^{T} \times {\hat{F}}_{s}{(h,w)}}
    {\left \| {{P}}_{c}^{i}\right \| 
     \left \| {\hat{F}}_{s}{(h,w)}\right \|}, 
    \end{split}
\end{equation}  
where ${\hat{F}}_{s}{(h,w)} \in {\mathbb{R}}^{1\times D}$ represents the feature vector of the feature maps ${\hat{F}}_{s}$ at position $(h,w)$ along channel dimension. 

To incorporate the knowledge represented by the similarity maps $S_s$ into the cluster prototypes $P_c$, we also adopt a one-dimensional convolution to implement the computation of ${\scriptsize\circled{D}}$, as illustrated in Fig.~\ref{fig:fore}(b) and (c). 
Specifically, the computation begins with the channel-wise generation of convolution filters. 
Given that the cluster prototypes $P_c$ are arranged as shown in Fig.~\ref{fig:fore}(b). 
We slice $P_c$ along the channel-dimension and obtain $D$ convolution vectors $\{K_i\}_{i=1}^{D}$ where $K_i\in {\mathbb{R}}^{1 \times N_s}$ contains cluster prototypes' semantic component on the {\it i-th} channel. 
After that, as done in Fig.~\ref{fig:fore}(c), we conduct one-dimensional convolution to obtain fused maps ${\bar{F}_s} \in {\mathbb{R}}^{D \times H \times W}$. 
The computation for the $i$-th map can be expressed as:  
\begin{equation}\label{eqn:cov1d}
    \small
    \begin{split} 
    {\bar{F}_s^i} = {\rm{conv1D}}\left(\phi\left({S}_s\right), K_i\right),   
    \end{split}
\end{equation}
where $\phi(\cdot)$ is softmax operation, $\phi({S}_s)$ stands for probability map, $K_i$ work as the convolution filter. 
In this end, to suppress introduced noise in the fusion step, we tailor the fused map ${\bar{F}_s}$ to high-fidelity foreground prototype $P_f$ by Mask Average Pooling.
\begin{equation}\label{eqn:map}
    \begin{split} 
    P_f = \frac{\sum_{h,w} {\bar{F}}_s(h,w) \odot m_s(h, w)}{\sum_{h,w} m_s(h,w)}, 
    \end{split}
\end{equation}
where $m_s$ is the given mask of the support image and resized to the same as ${\bar{F}}_s$. 
\begin{figure*}[t]
    \begin{center}
        \includegraphics[width=0.95\linewidth,angle=0]{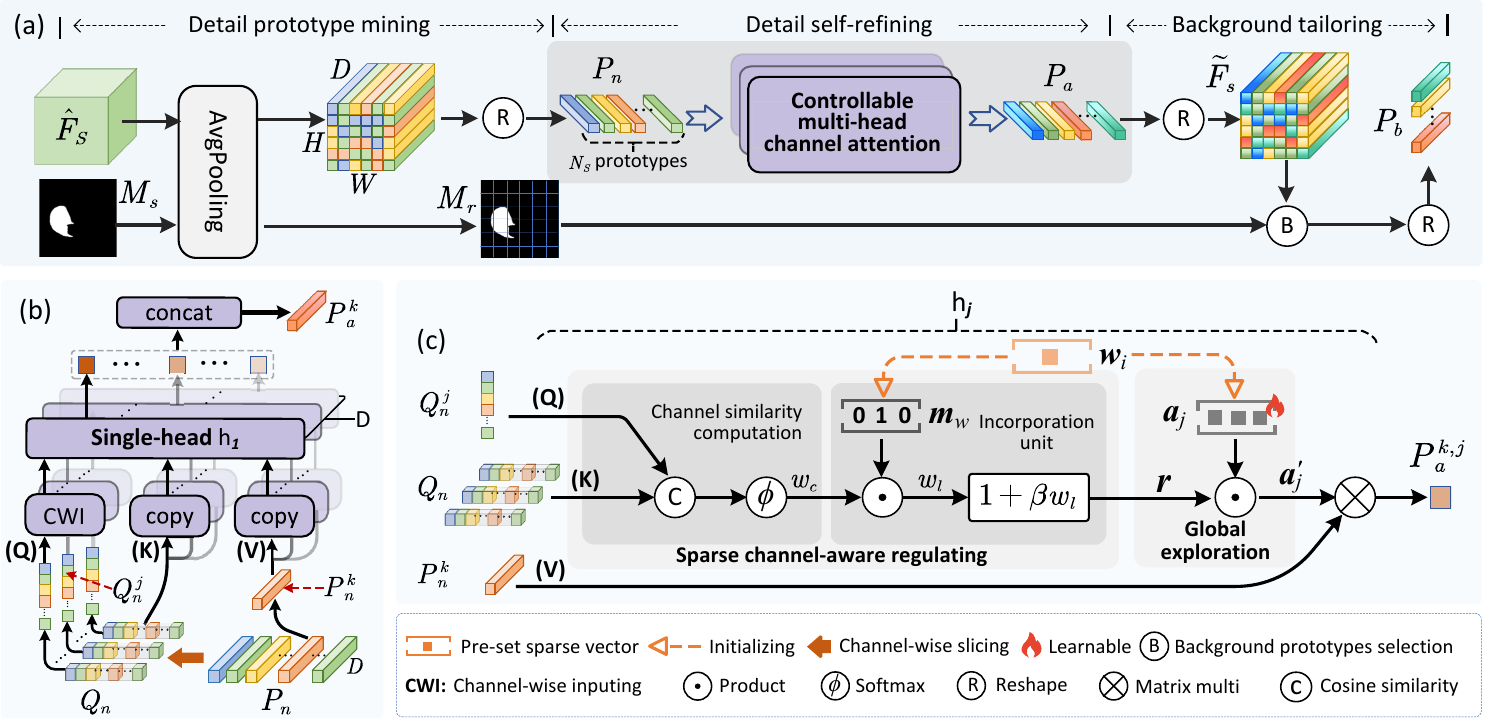}  
    \end{center} 
    \caption{
    Illustration of {\shortnameblna}. 
    {\bf (a)} shows the proposed {\shortnameblna} architecture where the {\it controllable multi-head channel attention} refreshes raw detail prototypes $P_n$ to $P_a$ by incorporating the channel-specific structural information. 
    Taking $P_n^k$ as an example, {\bf (b)} presents its detail self-refining to corresponding high-fidelity prototype $P_a^k$ whose $j$-th element $P_a^{k,j}$ is generated by attention head $h_j$.   
    {\bf (c)} elaborates $h_j$ where {\it sparse channel-aware regulation block} generate control factor ($\boldsymbol{r}$) to modulate global channel structural information of the $j$-th channel ($\boldsymbol{a}_j$) that is learnt by {\it global exploration block}.  
    }
    \label{fig:back}
\end{figure*}

{\bf Remark:} 
In Eq.~\eqref{eqn:cov1d}, $S_s$ is essentially semantic response maps concerning the cluster prototypes, such that probability map $\phi({S}_s)$ is noticeably relational to the detail semantics. 
That is, this fusion ensured by ${\scriptsize\circled{D}}$ computation is guided by the detail semantics represented in $S_s$. 
Equivalently, the fusion process preserves the detail semantics, as our expectation. 

Besides, two designs differs {\shortnamefspa} from the previous work. 
First, {\shortnamefspa} reduces the mined cluster prototypes to a fused one instead of using these prototypes separately~\citep{fan2022selfsup,liu2022intermediate}.
Second, the proposed channel-wise attention leads to global semantics preserving the local semantics unlike spatially weighting~\citep{li2021adaptive}. 


\subsection{Background Channel-structural Multi-head Attention}~\label{sec:blna}


Unlike the foreground taking the cluster prototypes as the local details, the background in medical images is usually semantic-less in a large scope.  
Therefore, in this paper, we do not mine from the spatial dimension but deem the structural information in the channel dimension as the local details. 
Within this context, we design a controllable channel attention mechanism to jointly model the channel-specific structural information and incorporate them into the raw background prototypes. 

{\bf Overview.} As illustrated in Fig.~\ref{fig:back}(a), {\shortnameblna} begins with generating raw detail prototypes. 
By Average Pooling and reshaping, ${\hat{F}}_{s}$ is converted to $P_n \in {\mathbb{R}}^{(H\times W) \times D}$. 
Following that, the \textit{controllable multi-head channel attention} module refreshes $P_n$ to high-fidelity background prototypes $P_a$. 
Finally, $P_a$ is reshaped to feature maps ${\widetilde{F}}_s$ and further tailored to the high-fidelity background prototypes $P_b$ by the background zone in pooled support mask ${M}_{r}$.

{\bf Controllable multi-head channel attention.} 
The proposed channel attention mechanism encodes the channel-structural information into raw background prototypes in an element manner. 
For a raw prototype, their elements are independently refined by the structural information of different channels. 
We achieve this by the D-way architecture illustrated in Fig.~\ref{fig:back}(b). 
Suppose that for any raw prototype in $P_n$, denoted by $P_n^k$, the converted high-fidelity prototype is $P_a^k$. 
In the {\bf Q}-{\bf K}-{\bf V} fashion, we set
${\rm{\bf Q}}\!=\!Q_n$, 
${\rm{\bf K}}\!=\!P_n$, 
${\rm{\bf V}}\!=\!P_n^k$ where $Q_n$ is transformed by channel-wise slicing $P_n$. 
In the proposed module, $P_n$ and $P_n^k$ are copied D times and inputted into the D heads, respectively. 
At the same time, $Q_n$ is inputted channel-wise. 
That is, the $j$-th head takes the $j$-th component $Q_n^{j}$ as the input.
The multi-head module refining $P_n^k$ can be formulated as
\begin{equation}\label{eqn:multi-head}
    \small
    \begin{split} 
    {{P}_a^k} &= {\rm{cat}}\left(\{P_a^{k,j}\}\right), \\
    P_a^{k,j} &= {\rm{h}}_j\left({Q}_n^{j}, Q_n, P_n^k \right), 1 \leq j \leq D,
    \end{split}
\end{equation} 
where ${\rm{cat}}(\cdot)$ concatenates the input set to a vector according to their indices; ${\rm{h}}_j(\cdot,\cdot,\cdot)$ is the $j$-th channel attention head generating $P_a^{k,j}$ (the $j$-th element in $P_a^k$) that we elaborate as follows.

In Eq.~\eqref{eqn:multi-head}, the objective of the attention head $h_j$ is encoding the $j$-th channel-specific structural information, denoted by $\boldsymbol{a}'_j$, into the raw $P_n^k$. 
Under the attention framework, the encoding can be implemented by a weighting operation $P_n^k \times \boldsymbol{a}'_j$, whilst the $\boldsymbol{a}'_j$  
generation is the core problem we need to address. 
For this issue, as depicted in Fig.~\ref{fig:back}(c), we provide a controllable design consisting of (i) global exploration module and (ii) sparse channel-aware regulating module. 
Among them, the former predicts the global channel structural information of the $j$-th channel $\boldsymbol{a}_j$, whilst the latter serves as a controller by injecting the $j$-th channel-specific adjustment $\boldsymbol{r}$.
Thus, the working mechanism ${\rm{h}}_j$ can be formulated as 
\begin{equation}\label{eqn:one-head}
    \begin{split} 
    {\rm{h}}_j = {P_n^k} \times \overbrace {\left(\boldsymbol{r} \odot \boldsymbol{a}_j \right)}^{\boldsymbol{a}'_j}, 
    \end{split}
\end{equation} 
where parameter ${\boldsymbol{a}_j}$ is learnable; operator $\odot$ means element-wise multiplying.

In Eq.~\eqref{eqn:one-head}, the generation of adjustment $\boldsymbol{r}$ involves two blocks in the sparse channel-aware regulating module (see the two dark grey box in Fig.~\ref{fig:back}(c)). 
First, the channel similarity computation, formulated by Eq.~\eqref{eqn:one-head-r-1}, captures the dynamics of the relationship between $j$-th channel and other channels. 
\begin{equation}\label{eqn:one-head-r-1}
    \begin{split}
    \boldsymbol{w}_c = \phi\left({\rm{cossim}}\left({Q_n},{Q_n^j}\right)\right),~~ 
    \boldsymbol{w}_{c,i}=\frac{\left({Q_n^i}\right)^{T} \times {Q_n^j}}
    {\left \| Q_n^i\right \| 
     \left \| Q_n^j\right \|},
    \end{split}
\end{equation}
where $\boldsymbol{w}_c \in {\mathbb{R}}^{D}$ is the channel similarity whose $i$-th element is $\boldsymbol{w}_{c,i}$, 
function ${\rm{cossim}}(\cdot,\cdot)$ measures the cosine similarity of vector ${Q_n^j}$ over set ${Q_n}$, $\phi$ is softmax operation. 
Subsequently, the incorporation unit generates adjustment coefficients by high-lighting the sparse-relative channels indexed by masked frozen vector ${\boldsymbol{m}_w}$.  
This process can be formulated as  
\begin{equation}\label{eqn:one-head-r-2}
    \begin{split}
    \boldsymbol{r} = 1 + \beta\left(\boldsymbol{w}_c \odot \boldsymbol{m}_w \right),
    \end{split}
\end{equation}
where rade-off parameter $\beta$ stands for the control strength.




As mentioned above, the proposed $h_j$ involves two important parameters, i.e., $\boldsymbol{a}_j$ (Eq.~\eqref{eqn:one-head}) and $\boldsymbol{m}_w$ (Eq.~\eqref{eqn:one-head-r-2}). 
In our design, both of them are initiated by a pre-set sparse vector $\boldsymbol{w}_i$ that represents a prior knowledge about the channel structural pattern. 
Specifically, at the beginning of model training, we set $\boldsymbol{m}_w={\rm{mask}}(\boldsymbol{w}_i)$ and $\boldsymbol{a}_j=\boldsymbol{w}_i$ where function ${\rm{mask}}(\cdot)$ outputs Boolean vector whose locations of 1 corresponds to the non-zero places in input vector.

{\bf Remark}: 
In our controllable attention mechanism, the core idea is imposing sparse channel-aware regulating to adjust the learnt global channel relation, leading to channel-specific structural information. Here, the sparse constraint is motivated by the ubiquitous sparse nature of neural connections, whose rationality is verified by much work~\citep{liu2015sparse,child2019generating}.  

Also, from a methodological point of view, our structure can be understood as a piece of work of structural learning-based attention~\citep{ramachandran2019stand,liu2021swin,hassani2023neighborhood}, but in the channel dimension. 
For instance, shifted window partitioning in Swin Transformer~\citep{liu2021swin} introduces spatial relation constraint to self-attention.
Similarly, our sparse channel-aware regulating introduces a channel structural constraint, i.e.,  sparse relation ($\boldsymbol{r}$), to the predicted 
global channel structural information ($\boldsymbol{a}_j$).   
\subsection{Loss Function} 
We regulate cross-entropy regularization to supervise this model training process: 
\begin{equation} \label{eqn:loss_mca}
    \small
    \begin{aligned}
    \mathcal{L}_{seg} \!=\!  
    - \frac{1}{HW}\displaystyle\sum_{h}^{H}\displaystyle\sum_{w}^{W}
    \displaystyle\sum_{j\in \{f,b\}}^{} {m}_{q}^{j}\left(h,w\right) \odot log\left(\hat{m}_{q}^{j}\left(h,w\right)\right),
    \end{aligned}
\end{equation}
where $\hat{m}_{q}^{j}(h,w)$ is the predicted results of the query mask label ${m}_{q}^{j}(h,w)$; in $\{f,b\}$, $f$ and $b$ means foreground and background, respectively. 
Also, following~\citep{wang2019panet,ouyang2020self,shen2023qnet}, we perform another inverse learning where the query images serve as the support set to predict labels of the support images.  
Thus, we encourage a prototypical alignment formulated by 
\begin{equation} \label{eqn:loss_mca} 
    \small
    \begin{aligned}
    \mathcal{L}_{reg} =  
    - \frac{1}{HW}\displaystyle\sum_{h}^{H}\displaystyle\sum_{w}^{W}
    \displaystyle\sum_{j\in \{f,b\}}^{} {m}_{s}^{j}(h,w) \odot log\left(\hat{m}_{s}^{j}(h,w)\right).
    \end{aligned}
\end{equation}

Overall, for each training episode, the final objective of {\shortmodelname} is defined as follows: 
\begin{equation}\label{eqn:loss_final}
    \begin{split} 
    \mathcal{L}_{\rm{\shortmodelname}}= \mathcal{L}_{seg} + \mathcal{L}_{reg}. 
    \end{split}
\end{equation}

\section{Experiments}\label{sec:rlt}
This part first introduce the experimental settings, followed by the segmentation results on three challenging benchmarks. 
The extensive model discussion is provided in this end.
\begin{table*}[t]
    \caption{Experiment results (in Dice score) on {\bf ABD-MRI} and {\bf ABD-CT}.  
    Numbers in bold and italics indicate the best and the second-best results, respectively.}
    \label{tab:rlt-ABD}
    \renewcommand\tabcolsep{3.6pt} 
    \renewcommand\arraystretch{1.0} 
    \footnotesize
    \centering
    \begin{tabular}{c|l|c|c|c|c|c|c|c|c|c|c}
        \hline
        \multirow{2}{*}{\textbf{Settings}} &\multirow{2}{*}{\textbf{Method}} &\multicolumn{5}{c|}{\textbf{ABD-MRI}} & \multicolumn{5}{c}{\textbf{ABD-CT}}\\ 
        \cline{3-12}
         & & \textbf{Liver} & \textbf{R.kidney} & \textbf{L.kidney} & \textbf{Spleen} & \textbf{Mean} & \textbf{Liver} & \textbf{R.kidney} & \textbf{L.kidney} & \textbf{Spleen} & \textbf{Mean} \\
        \hline
        \multirow{6}{*}{\makecell{Setting-1}} 
        & SE-Net~\citep{roy2020squeeze}          &{29.02}  &{47.96}  &{45.78}  &{47.30}  &{42.51} &{35.42} &{12.51} &{24.42} &{43.66} &{29.00}\\
        & PANet~\citep{wang2019panet}            &{47.37}  &{30.41}  &{34.96}  &{27.73}  &{35.11} &{60.86} &{50.42} &{56.52} &{55.72} &{57.88}\\
        & SSL-ALPNet~\citep{ouyang2020self}      &{70.49}  &{79.86}  &\textit{81.25}  &{64.49}  &{74.02} &{67.29} &{72.62} &{76.35} &\textbf{70.11} &{71.59}\\
        & Q-Net~\citep{shen2023qnet}             &\textit{73.54}  &\textit{84.41}  &{68.36}  &\textbf{76.69}  &\textit{75.75} &{68.65} &{55.63} &{69.39} &{56.82} &{62.63} \\
        & CAT-Net~\citep{lin2023few}             &{73.01}  &{79.54}  &{73.11}  &{69.31}  &{73.74} &{66.24} &{47.83} &{69.09} &{66.98} &{62.54}\\
        \rowcolor{gray! 40} & {\bf \shortmodelname}~(our)  &\textbf{75.06}  &\textbf{85.37}  &\textbf{81.88}  &\textit{70.93}  &\textbf{78.31} &\textbf{69.32} &\textbf{74.54} &\textbf{78.01} &{69.31} &\textbf{72.79}\\
        \hline
        \multirow{6}{*}{\makecell{Setting-2}}
        & SE-Net~\citep{roy2020squeeze}      &{27.43}  &{61.32}  &{62.11}  &{51.80}  &{50.66} &{0.27} &{14.34} &{32.83} &{0.23} &{11.91}\\ 
        & PANet~\citep{wang2019panet}        &{69.37}  &\textit{66.94}  &{63.17}  &{61.25}  &{65.68} &{61.71} &{34.69} &{37.58} &{43.73} &{44.42}\\ 
        & SSL-ALPNet~\citep{ouyang2020self}  &{69.46}  &{62.34}  &\textit{75.49}  &\textbf{69.02}  &{69.08} &{66.21} &\textbf{64.68} &{58.66} &\textbf{66.69} &{64.06}\\
        & Q-Net~\citep{shen2023qnet}         &\textbf{82.97}  &{51.81}  &{70.39}  &{57.74}  &\textit{65.73} &{64.44} &{41.75} &{66.21} &{37.87} &{52.57}\\ 
        & CAT-Net~\citep{lin2023few}         &{74.09}  &{63.51}  &{70.56}  &{67.02}  &{68.79} &{52.53} &{46.87} &{65.01} &{46.73} &{52.79} \\
        \rowcolor{gray! 40}   &{\bf \shortmodelname}~(our)  &\textit{78.56}  &\textbf{82.01}  &\textbf{76.47}  &\textit{68.27}  &\textbf{76.33} &\textbf{69.16} &{63.55} &\textbf{68.46} &{66.48} &\textbf{66.17} \\
        \hline
    \end{tabular}
\end{table*}
\begin{figure*}[t]
    \begin{center}
        \subfigure{
            \includegraphics[width=0.90\linewidth]{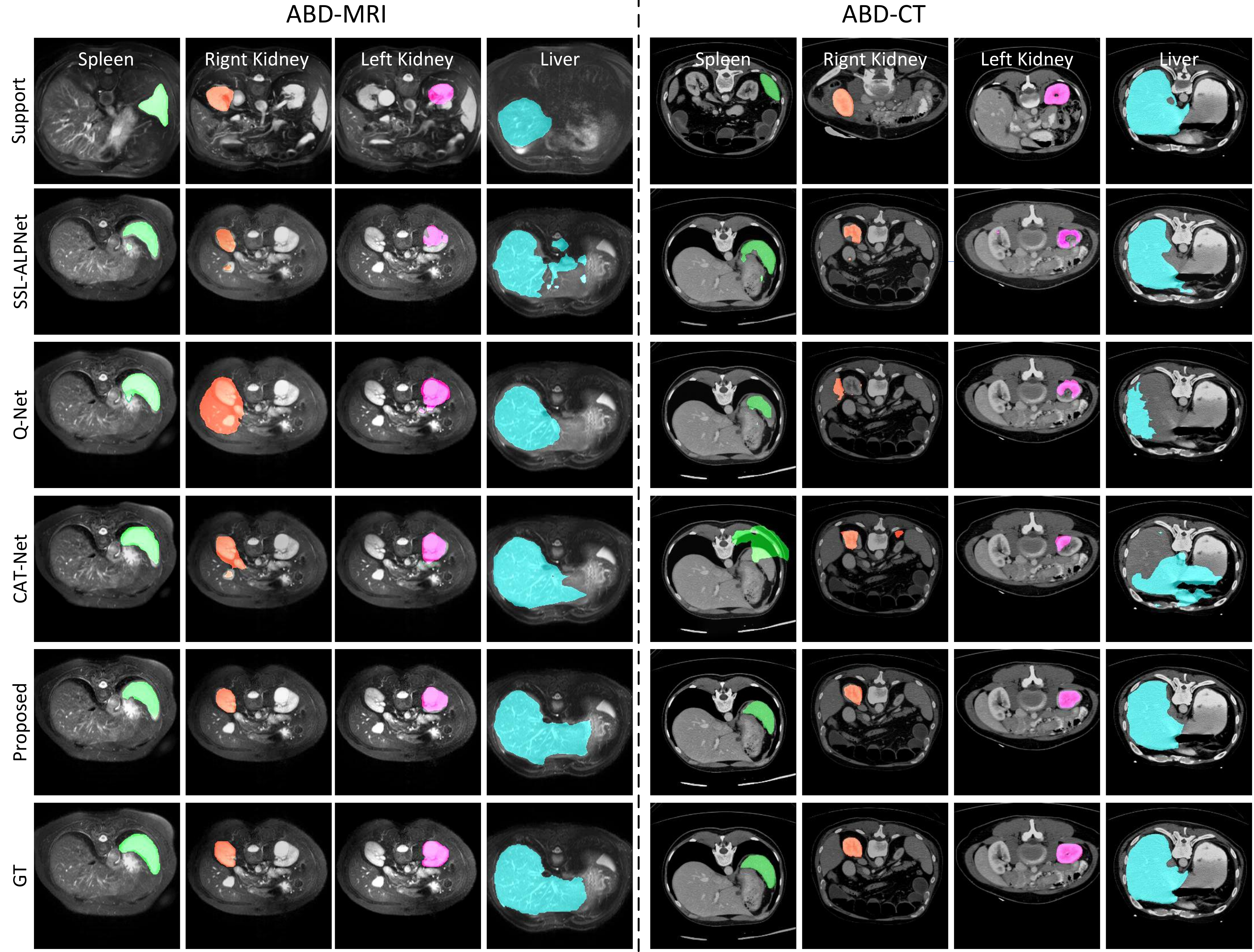}}
        \end{center}
        \setlength{\abovecaptionskip}{0cm}
        \caption{
        The qualitative comparison results in the {\bf ABD-MRI} dataset (the left side) and {\bf ABD-CT} dataset (the right side) under strict Setting-2. {\bf Top to bottom}: Support images, segmentation results and ground-truth segmentation of a query slice containing the target object (Best viewed with zoom). 
        } 
    \label{fig:vis-abd}
\end{figure*}
\begin{figure*}[!thp]
    \begin{center}
        \subfigure{
            \includegraphics[width=0.90\linewidth]{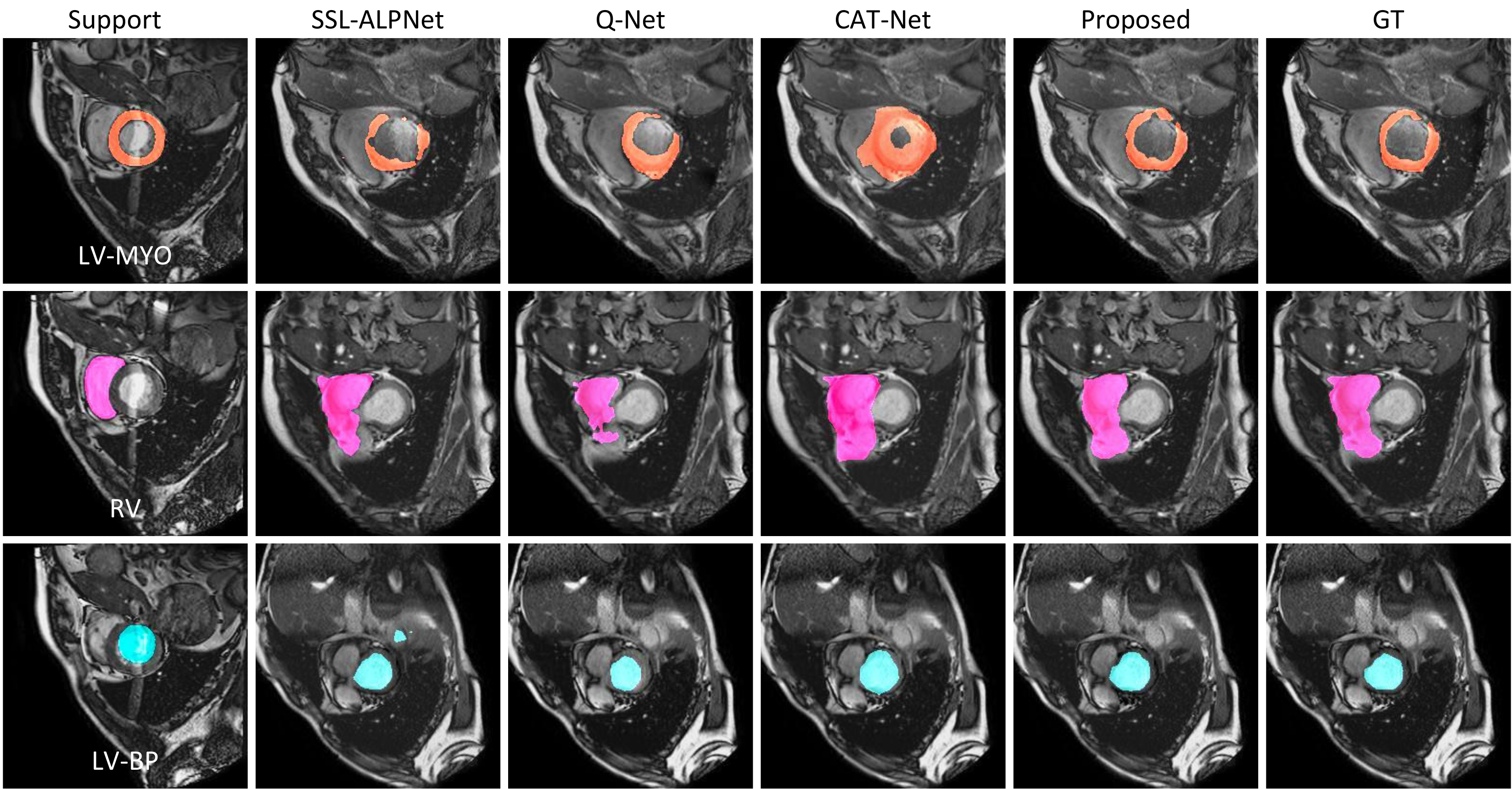}}
        \end{center}
        \setlength{\abovecaptionskip}{0cm}
        \caption{
        The qualitative comparison results in the {\bf CMR} dataset under Setting-1. 
        {\bf Left to right}: Support images, segmentation results and ground-truth segmentation of a query slice containing the target object. 
        {\bf Top to bottom}: LV-MYO (left ventricular myocardium), RV (right ventricle) and LV-BP (left ventricular outflow tract blood pool). (Best viewed with zoom) 
        } 
    \label{fig:vis-cmr}
\end{figure*}


\subsection{Data Sets}
To demonstrate the effectiveness of {\shortmodelname}, we conduct evaluation on three challenging datasets with different segmentation scenarios. Their details are presented as follows.


{\bf Abdominal CT dataset}~\citep{Landman2015chaos}, termed {\bf ABD-CT}, was acquired from the Multi-Atlas Abdomen Labeling challenge at the Medical Image Computing and Computer Assisted Intervention Society (MICCAI) in 2015. 
This dataset contains 30 3D abdominal CT scans. Of note, this is a clinical dataset containing patients with various pathologies and variations in intensity distributions between scans.

{\bf Abdominal MRI dataset}~\citep{kavur2021chaos}, termed {\bf ABD-MRI}, was obtained from the Combined Healthy Abdominal Organ Segmentation (CHAOS) challenge held at the IEEE International Symposium on Biomedical Imaging (ISBI) in 2019. 
This dataset consists of 20 3D MRI scans with a total of four different labels representing different abdominal organs.

{\bf Cardiac MRI dataset}~\citep{zhuang2018multivariate}, termed {\bf CMR}, was obtained from the Automatic Cardiac Chamber and Myocardium Segmentation Challenge held at the Conference on Medical Image Computing and Computer Assisted Intervention (MICCAI) in 2019. 
It contains 35 clinical 3D cardiac MRI scans.

In our experiment settings, to ensure fair comparison, we adopted the same image preprocessing solution as SSL-ALPNet~\citep{ouyang2020self}. 
Specifically, we sampled the images into slices along the channel dimension, and resized each slice to 256$\times$256 pixels. Moreover, we repeated each slice three times along the channel dimension to fit into the network. 
We employ 5-fold cross-validation as our evaluation method, where each dataset is evenly divided into 5 parts.


\subsection{Evaluation Protocol}
To evaluate the performance of the segmentation model, we utilized the conventional Dice score scheme. 
The Dice score has a range from 0 to 100, where 0 represents a complete mismatch between the prediction and ground truth, while 100 signifies a perfect match. The Dice calculation formula is
\begin{equation}\label{eqn:loss_mca}
\begin{split} 
    Dice(A,B)=\frac{2\begin{Vmatrix}
    A\cap B
    \end{Vmatrix}}{\begin{Vmatrix}
    A
    \end{Vmatrix}+ \begin{Vmatrix}
    B
    \end{Vmatrix}} \times 100\%, 
\end{split}
\end{equation}
where $A$ represents the predicted mask and $B$ represents the ground truth. 

\begin{table*}[!thp]
    \caption{Experiment results (in Dice score) on the {\bf CMR} dataset. 
    Numbers in bold and italics indicate the best and the second-best results, respectively.}
    \label{tab:rlt-CMR} 
    \renewcommand\tabcolsep{20pt} 
    \renewcommand\arraystretch{1.0}
    \footnotesize
    \centering 
    \begin{tabular}{c|l|c|c|c|c}
        \hline
        \textbf{Settings} &\textbf{Method} &\textbf{RV} &\textbf{LV-MYO} &\textbf{LV-BP} 
        &\multicolumn{1}{c}{\textbf{Mean}} \\ 
        \hline
        \multirow{5}{*}{\makecell{Setting-1}} 
        & SE-Net~\citep{roy2020squeeze}      &{12.86}  &{58.04}  &{25.18}   &{32.03} \\
        & PANet~\citep{wang2019panet}        &{57.13}  &\textbf{72.77}  &{44.76}   &{58.20} \\
        & SSL-ALPNet~\citep{ouyang2020self}  &\textit{77.59}  &{63.29}  &{85.36}   &\textit{75.41} \\
        & Q-Net~\citep{shen2023qnet}         &{67.99}  &{52.09}  &\textit{86.21}   &{68.76}  \\
        & CAT-Net~\citep{lin2023few}                   &{69.37}  &{48.81}  &{81.33}   &{66.51} \\
        \rowcolor{gray! 40}   &{\bf \shortmodelname}~(our) &\textbf{79.73}  &\textit{64.91}  &\textbf{87.75}  & \multicolumn{1}{c}{\textbf{77.46}}  \\
        \hline
    \end{tabular}
\end{table*}

\begin{table*}[t]
    \caption{Ablation study results on the {\bf ABD-MRI} dataset. 
    Numbers in bold indicate the best results.} 
    \label{tab:ABLA}
    \renewcommand\tabcolsep{5pt} 
    \renewcommand\arraystretch{1.1}
    \footnotesize
    \centering
    \begin{tabular}{l l |c c c| c c c c c | c c c c c}
        \hline
        \multirow{2}{*}{\textbf{\#}} &\multirow{2}{*}{\textbf{Method}} &\multirow{2}{*}{RAN} &\multirow{2}{*}{\shortnamefspa} &\multirow{2}{*}{\shortnameblna} &\multicolumn{5}{c|}{\textbf{Setting-1}}  &\multicolumn{5}{c}{\textbf{Setting-2}} \\
        & & & & &{Liver} &{R.kidney} &{L.kidney} &{Spleen} &{Mean} &{Liver} &{R.kidney} &{L.kidney} &{Spleen} &{Mean}  \\ 
        \hline
        1 &SSL-ALPNet      &-- &-- &-- &{70.49}  &{79.86} &{81.25} &{64.49} &{74.02} &{69.46}  &{62.34}  &{75.49}  &\textbf{69.02}  &{69.08} \\
        \hline
        2 &{\shortmodelname} w/o RAN    &\xmark &\cmark &\cmark &{72.92}  &{81.98} &\textbf{85.55} &{65.66} &{76.52} &{70.54}  &{72.97} &\textbf{80.94} &{63.53} &{71.99} \\
        3 &{\shortmodelname} w/o \shortnamefspa  &\cmark &\xmark &\cmark &{71.97}  &{82.14} &{80.95} &{67.46} &{75.63} &{74.86}  &{73.56} &{72.97} &{66.99} &{72.09} \\
        4 &{\shortmodelname} w/o \shortnameblna  &\cmark &\cmark &\xmark &{70.24}  &{84.64} &{80.81} &{68.04} &{75.93} &{66.09}  &{73.47} &{77.19} &{66.22} &{70.74} \\
        \hline
        5 &{\shortmodelname} w/ RAN   &\cmark &\xmark &\xmark &{75.89}  &{80.89} &{77.07} &{66.73} &{75.15} &{74.86}  &{68.69} &{70.89} &{60.47} &{68.73} \\
        \rowcolor{gray! 40} 6 &{\shortmodelname} &\cmark &\cmark &\cmark &\textbf{75.06}  &\textbf{85.36} &{81.88} &\textbf{70.93} &\textbf{78.31} &\textbf{78.56}  &\textbf{82.01} &{76.47} &{68.27} &\textbf{76.33} \\ 
        \hline
    \end{tabular}
\end{table*}

\begin{figure}[t]
    \begin{center}
        \subfigure{
            \includegraphics[width=0.97\linewidth]{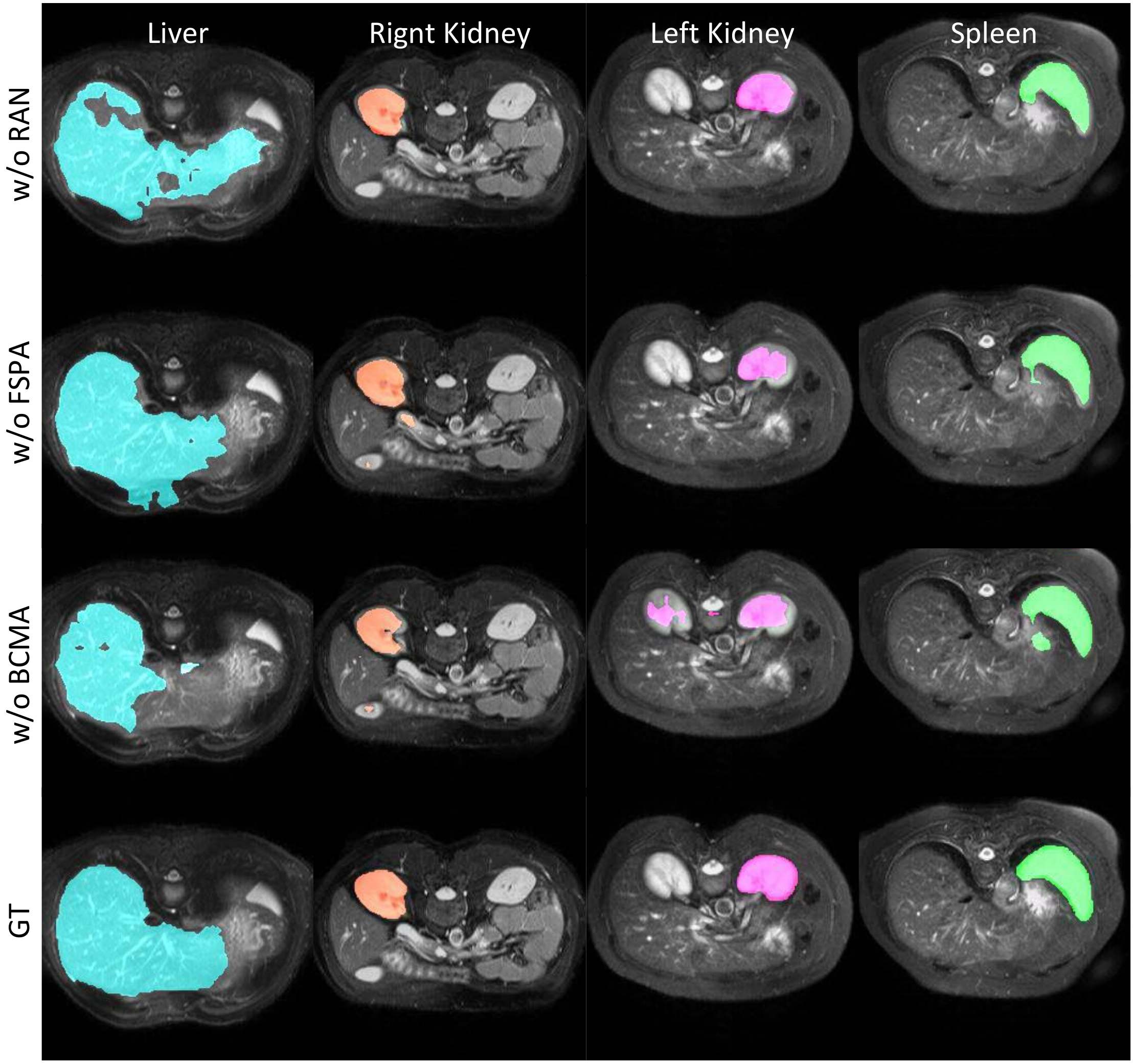}}
        \end{center}
        \setlength{\abovecaptionskip}{0cm}
        \caption{
        The qualitative comparison results of ablation study in the {\bf ABD} dataset under Setting-2. 
        {\bf Left to right}: Liver, Right kidney, Left kidney and Spleen. (Best viewed with zoom)
        } 
    \label{fig:vis-abla}
\end{figure}


\subsection{Few-Shot Settings}
To evaluate the model's performance, we follow the experimental settings in~\citep{ouyang2020self,hansen2022anomaly}, considering two cases. 
{\bf Setting-1} is the initial setting proposed in~\citep{roy2020squeeze}, where test classes may appear in the background of training images. 
We train and test on all classes in the dataset without any partitioning.  
{\bf Setting-2} is a strict version of Setting-1, proposed in~\citep{ouyang2020self}, where we adopted a stricter approach. 
In this setting, test classes do not appear in any training images. 
For instance, when segmenting Liver during training, the support and query images do not contain the Spleen, which is the segmenting target for testing.  
We directly removed the images containing test classes during the training phase to ensure that the test classes are truly "unseen" for the model.

\subsection{Implementation Details} \label{sec:imp-detail}
We implemented our model using the Pytorch framework with a pre-trained fully convolutional Resnet101 model as the feature extractor. 
The Resnet-101 model was pre-trained on the MS-COCO dataset. 
Given that the superpixel pseudo-labels contain rich clustering information, which are helpful to alleviate the annotation absence.
We generate the superpixel pseudo-label in an offline manner as the support image mask before starting the model training, following~\citep{ouyang2020self,hansen2022anomaly,shen2023qnet}.

\changed{In {\shortmodelname}, there is one hyper-parameter: Local adjustment intensity $\alpha$ in Eq.~\eqref{eqn:one-head}.
As another important factor, the sparse pattern of $\boldsymbol{w}_i$ follows the neighbour channel constraint, namely, $\boldsymbol{w}_i = \left[ \boldsymbol{0}, w_1, w_2, w_3, \boldsymbol{0} \right]$    
where $w_2$ is the $j$-th element of $\boldsymbol{w}_i$, $({w}_{1},{w}_{2},{w}_{3}) \! \in \! \left[ 0, 1.0 \right]$, ${w}_{2} \!>\! {w}_{1}={w}_{3}$.}   
Specifically, in the {\bf ABD-MRI} dataset, Setting-1 adopts $\beta\!=\!0.3, (w_1,w_2,w_3)\!=\!(0.2,0.8,0.2)$, whilst $\beta\!=\!0.2, (w_1,w_2,w_3)\!=\!(0.3,0.6,0.3)$ are used in Setting-2. 
In the {\bf ABD-CT} dataset, Setting-1 adopts $\beta\!=\!0.2, (w_1,w_2,w_3)\!=\!(0.3,0.7,0.3)$, and $\beta\!=\!0.4, (w_1,w_2,w_3)\!=\!(0.1,0.7,0.1)$ are used in Setting-2. 
For the {\bf CMR} dataset, Setting-1 selects $\beta\!=\!0.3, (w_1,w_2,w_3)\!=\!(0.1,0.9,0.1)$.

For our experimental results, we used stochastic gradient descent algorithm with a batch size of 1 for 100k iterations to minimize the objective in Eq.~\eqref{eqn:loss_final}. 
The self-supervised training took around 4.5 hours on a single Nvidia TITAN V GPU, and the memory consumption was about 8.1GB. 

\subsection{Competitors}
To evaluate our approach, we compared it with six state-of-the-art medical image semantic segmentation methods, 
including 
SE-Net~\citep{roy2020squeeze}, 
PANet~\citep{wang2019panet}, 
SSL-ALPNet~\citep{ouyang2020self},
Q-Net~\citep{shen2023qnet},
and CAT-Net~\citep{lin2023few}. 
Among them, SE-Net belongs to the category of constructing support images-based guidance, whilst the rest comparisons all follow the clue of prototypic network.  
For a fair comparison, we obtain the results of all prototype-based methods, i.e., PANet, SSL-ALPNet, Q-Net and CAT-Net, by re-running their official codes on the same evaluating bed with {\shortmodelname}.  
The results of SE-Net are cited from the publication.

\subsection{Quantitative and Qualitative Results} 
The same as the previous methods, we perform the evaluation on {\bf ABD-MRI} and {\bf ABD-CT} under both {Setting-1} and {Setting-2} whilst the {\bf CMR} is based on {Setting-1}. 
Tab.~\ref{tab:rlt-ABD} reports the results in Dice score on {\bf ABD-MRI} and {\bf ABD-CT}. 
The results showed that {\shortmodelname} outperforms the previous methods in the two settings.
On {\bf ABD-MRI} dataset, compared with the second-best method Q-Net in mean score, {\shortmodelname} achieves an improvement of {\bf 2.6} under Setting-1. 
Meanwhile, as for the strict Setting-2 testing model for "unknown" classes, {\shortmodelname} demonstrates impressive performance with {\bf 7.8} increase, especially with a dice score of approximately {\bf 82} for Right kidney. 
The reason is discussed in \texttt{Section~\ref{sec:abla}: Ablation study}.
On the {\bf ABD-CT} dataset, in average score, {\shortmodelname} also surpasses the second-best method SSL-ALPNet by {\bf 1.2} in Setting-1 and {\bf 2.1} in Setting-2, respectively. 
For an intuitive observation, we present the visual segmentation results in Fig.~\ref{fig:vis-abd}. 
As shown in this figure, {\shortmodelname} has much better segmentation for large objects (see Liver), while predicting the finer boundary for small objects (see Spleen).

Tab.~\ref{tab:rlt-CMR} shows the comparison results on {\bf CMR} with adjacent organs. 
In this scenario, {\shortmodelname} exhibited better segmenting performance in all three classes, obtaining {\bf 2.0} improvement in mean score compared with the previous best method SSL-ALPNet. 
The right side of Fig.~\ref{fig:vis-abd} depicts three toy experimental results. 
It is seen that the {\shortmodelname} can generate complicated boundaries (see LV-MYO, RV), implying more details are captured by {\shortmodelname} compared with the previous methods. 
For the objects with relatively regular shapes, e.g., LV-BP, {\shortmodelname} achieves fuller segmentation near the boundary.

\begin{table}[t]
    \caption{
    Effect analysis of channel-wise fusion in {\shortnamefspa} on the {\bf ABD-MRI} dataset under Setting-2. 
    Numbers in bold indicate the best results.
    }
    \label{tab:ana-flpa}
    \renewcommand\tabcolsep{4.5pt}
    \renewcommand\arraystretch{1.1}
    \footnotesize
    \centering
    \begin{tabular}{l|ccccc}
        \hline
        \textbf{Method}  &{Liver}  &{R.kidney}  &{L.kidney}  &{Spleen} &{Mean} \\ 
        \hline
        SSL-ALPNet       &{69.46}  &{62.34}  &{75.49}  &{\bf 69.02}  &{69.08}   \\
        \hline
        {\shortmodelname}-F-separating          &{72.13}  &{79.64}  &{73.21}  &{65.31}  &{72.56}  \\
        {\shortmodelname}-F-weighting           &{69.14}  &{65.97}  &{64.77}  &{59.89}  &{64.94}  \\
        \rowcolor{gray! 40} {\shortmodelname}   &\textbf{78.56}  &\textbf{82.01}  &\textbf{76.47}  &{68.27}  &\textbf{76.33}   \\ 
        \hline
    \end{tabular}
\end{table}
\begin{table*}[t]
    \caption{Effect analysis results for components in the sparse channel-aware regulation on {\bf ABD-MRI} under Setting-2. 
    SparseIni means that the attention matrix $\boldsymbol{a}$ is initiated by the sparse channel-aware $\boldsymbol{w}_i$.
    Numbers in bold indicate the best results.}
    \label{tab:blna-ana}
    \renewcommand\tabcolsep{11pt}
    \renewcommand\arraystretch{1.1}
    \footnotesize
    \centering
    \begin{tabular}{l l|ccc|ccccc}
        \hline
        \# &\textbf{Method}  &Learnable   &Adjust   &SparseIni   &{Liver}  &{R.kidney}  &{L.kidney}  &{Spleen} &{Mean} \\ 
        \hline
        1 &SSL-ALPNet               &-- &-- &-- &{69.46}  &{62.34}  &{75.49}  &{69.02}  &{69.08}  \\
        \hline
        2 &{\shortmodelname} w/o NCR                       &-- &-- &-- &{68.85}  &{78.01}  &{73.43}  &{62.23}  &{70.63} \\
        \hline
        3 &{\shortmodelname} w/ $\boldsymbol{a}$-fix        &\xmark &\cmark &\cmark &{71.99}  &{73.55}  &\textbf{77.48}  &\textbf{69.06}  &{73.02}  \\
        4 &{\shortmodelname} w/ $\boldsymbol{a}$-no-adjust  &\cmark &\xmark &\cmark &{75.41}  &{75.39}  &{72.23}  &{75.16}  &{74.55}  \\
        5 &{\shortmodelname} w/ $\boldsymbol{a}$-random     &\cmark &\cmark &\xmark &{45.77}  &{38.99}  &{39.42}  &{40.69}  &{41.22}  \\ 
        \rowcolor{gray! 40} 6 &{\shortmodelname}            &\cmark &\cmark &\cmark &\textbf{78.56}  &\textbf{82.01}  &{76.47}  &{68.27}  &\textbf{76.33} \\ 
        \hline
    \end{tabular}
\end{table*}

\subsection{Alation Study} \label{sec:abla}
As illustrated in the middle of Fig.~\ref{fig:fw}, {\shortmodelname} involves three components, i.e., RAN, {\shortnamefspa} and {\shortnameblna}. 
In this part, we carry out an ablation study to isolate their effect as follows. 
All experimental results are obtained based on the {\bf ABD-MRI} dataset under strict  Setting-2. 

\subsubsection{Effect to final performance}
By removing the three ones from our framework, we have variation methods: 
\begin{itemize}
    \item {\bf {\shortmodelname} w/o RAN}. We remove the RAN block and set the fused support feature ${\hat{F}}_s=F_s$ directly.      
    \item {\bf {\shortmodelname} w/o \shortnamefspa}. When {\shortnamefspa} block is removed, we generate the foreground prototype $P_f$ exploiting the conventional MAP skill, the same as previous work~\citep{ouyang2020self}.    
    \item {\bf {\shortmodelname} w/o \shortnameblna}. After removing the {\shortnameblna} block, the background prototypes $P_b$ is generated in two steps: (i) We convert the fused support feature ${\hat{F}}_{s}$ to feature maps by AP and then (ii) directly tailored to $P_b$ according to the background zone in support mask ${M}_{r}$, which is also generated by Average Pooling.  
\end{itemize}

From the results from Tab.~\ref{tab:ABLA}, we see that when removing any one of the three, the mean results have decline to some extent compared with {\shortmodelname}, whilst all being better than SSL-ALPNet. 
These results confirm that the proposed three designs all play positive roles in the proposed scheme. 
Meanwhile, the full version, {\shortmodelname}, significantly outperforms the other three variation methods. 
The results indicate that the three designs jointly lead to the final performance. 

To better understand the effect of the three designs, we present some typical segmentation results under Setting-2, as shown in Fig.~\ref{fig:vis-abla}. 
When any one is unavailable, the segmentation has evident 
deterioration. 
For example, when RAN is unavailable, the big object segmentation will have obvious holes (see Liver).
Due to removing background-specific {\shortnameblna}, some background zones are wrongly segmented, as adopting {\shortmodelname} w/o {\shortnameblna} (see Left Kidney, Spleen).


Combining results of Setting-1 with Setting-2, 
we have one detailed finding. 
First, {{\shortmodelname} w/o \shortnamefspa}, {{\shortmodelname} w/o \shortnameblna} have similar results with especially tiny gap under Setting-1, implying their balanced effect.
Unlike it, under Setting-2, {{\shortmodelname} w/o \shortnameblna} beat SSL-ALPNet by increase of {\bf 1.9} only, but has {\bf 3.1} decrease compared with {{\shortmodelname} w/o \shortnamefspa}. 
The comparison shows that for the truly ”unseen” scenario, background-oriented {\shortnameblna} is more important than foreground-oriented {\shortnamefspa}. 
The result is understandable: Performing detail self-refining on the background prototype is the most logical strategy when these training images cannot provide valuable references for the unseen testing classes.  
This is because, under Setting-2, {\shortmodelname} has a large performance margin on top of the previous methods (see Tab.~\ref{tab:ABLA}), which lose focus on ameliorating the background prototypes.

\subsubsection{Effect of RAN to {\shortnamefspa} and {\shortnameblna}}
As shown in Fig.~\ref{fig:fw}, the working of {\shortnamefspa} and {\shortnameblna} builds on RAN. 
Here, we propose another variation method of {\shortmodelname}, named {\shortmodelname} w/ RAN, to determine its effect.
In this comparison method, both {\shortnamefspa} and {\shortnameblna} are removed: The foreground class prototype and background detail prototypes are generated by traditional MAP and AP, respectively.  
As listed in Tab.~\ref{tab:ABLA} (see the fifth row), {{\shortmodelname} w/ RAN} improve by only {\bf 1.1} under Setting-1 and has a tiny gap of {\bf 0.3} under Setting-2, compared with SSL-ALPNet. 
This result indicates that RAN cannot work alone and must work jointly with {\shortnamefspa} and {\shortnameblna}.

\subsection{Model Analysis} \label{sec:ana-blna}

\subsubsection{Analysis of {\shortnamefspa}.}
This part discusses the two key features of {\shortnameblna}: (i) fusing the mined cluster prototypes into a single one for incorporating the local and global semantics and (ii) the channel-wise fusion strategy instead of weighting prototypes. 
To evaluate their effects, we propose two variations of {\shortmodelname}: 
\begin{itemize}
    \item {\bf {\shortmodelname}-F-separating}: Feature map ${\bar{F}}_s$ in Fig.~\ref{fig:fore}(a) is generated by directly computing cosine distance between the cluster prototypes $P_c$ and semantics fused feature ${\hat{F}}_s$. 
    \item {\bf {\shortmodelname}-F-weighting}: We average the cluster prototypes $P_c$ and employ the weighted prototype to compute cosine distance with ${\hat{F}}_s$. 
\end{itemize}
As listed in Tab.~\ref{tab:ana-flpa}, {{\shortmodelname}-F-separating} is {\bf 3.77} lower than {\shortmodelname} in the mean score and outperforms SSL-ALPNet by {\bf 3.48}. 
This comparison indicates that mining cluster prototypes can boost the segmentation but suffering from the loss of global semantics. 
This is in line with our expectations.
Besides, {{\shortmodelname}-F-weighting} is defeated by {\shortmodelname} with a large decrease of {\bf 11.39}, even lower than SSL-ALPNet. 
The results show that the weighting scheme will confuse the semantics and our channel-wise fusion provides a potential semantics incorporation way from local to global.

\subsubsection{Analysis of {\shortnameblna}} 
As shown in Fig.~\ref{fig:back}(b), the sparse channel-aware regulating is the core difference from the conventional channel attention mechanism. 
Evaluation in this part first focuses on the effect of this regulation.
To this end, we propose a comparison method, named {\bf {\shortmodelname} w/o NCR}, where the inputted prototype is directly refreshed by the channel similarity vector.  
From the second row in Tab.~\ref{tab:blna-ana}, we can see that {\shortmodelname} w/o NCR lowers {\shortmodelname}  by {\bf 5.7} and very close to result of removing {\shortnameblna}, i.e., {\shortmodelname} w/o BCMA (see Tab.~\ref{tab:ABLA}). 
The comparison indicates that the effect of {\shortnameblna} almost derives from our design of neighbour channel-aware regulation, confirming the rationality of introducing channel structural information.

As mentioned in \texttt{Section~\ref{sec:blna}}, the sparse channel-aware regulation contains three significant designs: 
(i) $\boldsymbol{a}$ is learnable, 
(ii) incorporation unit integrates $\boldsymbol{r}$ to adjust $\boldsymbol{a}$, and 
(iii) $\boldsymbol{a}$ is initiated by the sparse vector $\boldsymbol{w}_i$ representing the neighbour channel constraint.   
To demonstrate their effectiveness, we conduct a comparison experiment where three variation methods of {\shortmodelname} are given:  
\begin{itemize}
    \item {\bf {\shortmodelname} w/ $\boldsymbol{a}$-fix}: We keep $\boldsymbol{a}=\boldsymbol{w}_i$ during training. 
    
    \item {\bf {\shortmodelname} w/ $\boldsymbol{a}$-no-adjust}: Setting $\beta=0$ removes $\boldsymbol{r}$'s adjustment, whilst $\boldsymbol{a}$ is still learnable and initiated by $\boldsymbol{w}_i$. 
    
    \item {\bf {\shortmodelname} w/ $\boldsymbol{a}$-random}: $\boldsymbol{a}$ is not initiated by $\boldsymbol{w}_i$, instead, using conventional random initiation. 
\end{itemize}

\begin{figure}[t]
    \begin{center}
        \subfigure{
            \includegraphics[width=0.95\linewidth]{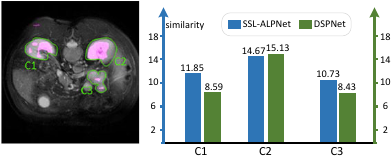}}
        \end{center}
        \setlength{\abovecaptionskip}{-0.3cm}
        \caption{Quality comparison between conventional prototypes and high-fidelity prototypes. 
        {\bf Left:} The testing image from {\bf ABD-MRI}. C1$\sim$C3 are the selected zones for comparison. 
        The segmentation in pink is the result of SSL-ALPNet.  
        {\bf Right:} Similarity score comparison of SSL-ALPNet and {\shortmodelname}.
        } 
    \label{fig:sg-ana}
\end{figure}


From the comparison results in Tab.~\ref{tab:blna-ana}, we have three main observations.  
{\it First}, {\shortmodelname}'s minimal version {{\shortmodelname} w/ $\boldsymbol{a}$-fix} is surpassing SSL-ALPNet by {\bf 5.97} in mean accuracy.  
This indicates that our neighbourhood-ware idea is effective, even when it works alone. 
Meanwhile, {\shortmodelname} surpasses {{\shortmodelname} w/ $\boldsymbol{a}$-fix} by {\bf 3.3}, indicating the importance of global fusion design ensured by enabling $A$ learnable. 
{\it Second}, {\shortmodelname} outperforms {{\shortmodelname} w/ $\boldsymbol{a}$-no-adjust} by {\bf 1.78} in mean score, confirming the rationality of introducing local adjusting. 
{\it Third}, compared with {\shortmodelname}, {{\shortmodelname} w/ $\boldsymbol{a}$-random}'s performance decrease sharply by {\bf 35.11}.
This result shows that the design of $\boldsymbol{w}_i$ initiation is crucial to optimising $\boldsymbol{a}$, once again supporting the importance of introducing the neighbourhood prior. 
Easily understood, $\boldsymbol{w}_i$ provides a good optimization initiation point.

\subsubsection{Conventional prototypes v.s. high-fidelity prototypes} 
Compared to the conventional prototypes, the core advantage of our prototypes is deeply representing the details. 
To verify it, we perform a quantitative experiment based on a typical image from the {\bf ABD} dataset. 
As shown in the left side of Fig.~\ref{fig:sg-ana}, we mark three zones containing objects, denoted by C1 (left kidney), C2 (right kidney) and C3 (gallbladder) in this image where C2 is the foreground.  
After that, we compute their similarity score under Setting-2 by averaging the final similarity map (i.e., ${\rm{cossim}}({F}_q,{P}_{k})$) at the locations of C1, C2 and C3.  
The right side in Fig.~\ref{fig:sg-ana} demonstrates the comparison results of {\shortmodelname} and SSL-ALPNet. 
Compared with SSL-ALPNet, {\shortmodelname} improved by {\bf 0.46} at C2. 
To the opposite, {\shortmodelname} declines by {\bf 3.26} and {\bf 2.3} at C1 and C3, respectively. 
In the view of max relative declination, e.g., $(S_{C2} - max(S_{C1},S_{C3}))/ S_{C2} \times 100\%$, SSL-ALPNet is {\bf 26.8}\%, whilst {\shortmodelname} amplify it to {\bf 44.3}\%.  
The results indicate that our high-fidelity prototypes can encourage more discriminative representations than the conventional prototypes. 


\begin{table}[t]
    \caption{
    Results of Dice score changing as parameter $\beta$ varying from 0.17 to 0.23 with step 0.01 (on the {\bf ABD-MRI} dataset under Setting-2).  
    }
    \label{tab:para-sensi}
    \renewcommand\tabcolsep{8.0pt}
    \renewcommand\arraystretch{1.0}
    \footnotesize
    \centering
    \begin{tabular}{ l | c c c c c }
        \hline
        $\beta$  &{Liver}  &{R.kidney}  &{L.kidney}  &{Spleen} &{Mean} \\ 
        \hline
        {0.18}   &{73.32}  &{79.67}  &{78.41}  &{68.95}  &{74.09}  \\
        {0.19}   &{67.58}  &{75.79}  &{73.41}  &{69.02}  &{75.45}  \\ 
        {0.20}   &{78.56}  &{82.01}  &{76.47}  &{68.27}  &{76.33}  \\ 
        {0.21}   &{73.98}  &{81.39}  &{72.64}  &{65.51}  &{73.38}  \\
        {0.22}   &{73.69}  &{81.13}  &{77.74}  &{65.13}  &{74.42}  \\
        {0.23}   &{72.01}  &{74.35}  &{76.36}  &{67.03}  &{72.44}  \\
        \hline
    \end{tabular}
\end{table}

\subsubsection{Parameter sensitiveness.} 
This part displays the performance sensitivity of the local adjustment intensity in Eq.~\eqref{eqn:one-head} based on the Setting-2 in the {\bf ABD} dataset.   
As presented in Tab.~\ref{tab:para-sensi}, when the parameter changes, there are no evident drops in the accuracy variation curves. 
This indicates that {\shortmodelname} is insensitive to the parameter $\beta$.

\section{Conclusion}\label{sec:con}
In this paper, we present a novel FSS approach, dubbed as {\shortmodelname}, aiming at the local information loss problem in medical images as adopting the prototypical paradigm. 
To our knowledge, this is an initial effort from the perspective: Enhancing detail representation ability of the off-the-shelf prototypes by detail self-refining. 
Specifically, we introduce two pivotal designs: FSPA and BLNA modules for the foreground class prototype and background detail prototypes generation, respectively. 
Among them, the former implements the detail self-refining by fusing the detailed prototypes clustered from the foreground.
The latter models this self-refining as incorporating the channel-specific structural information, employing the multi-head channel attention with sparse channel-aware regulation. 
{\shortmodelname}’s effectiveness is validated by state-of-the-art experimental results across three challenging datasets.

\section*{Declaration of Competing Interest}
The authors declare that they have no known competing financial interests or personal relationships that could have appeared to influence the work reported in this paper.


\subsection*{Acknowledgments}
This work is partly funded by the German Research Foundation (DFG) and National Natural Science Foundation of China (NSFC) in project Crossmodal Learning under contract Sonderforschungsbereich Transregio 169, the Hamburg Landesforschungsf{\"o}rderungsprojekt Cross, NSFC (61773083); NSFC (62206168, 62276048, 52375035).





\bibliographystyle{model2-names.bst}\biboptions{authoryear}
\bibliography{cas-refs-comm-ieee}



\end{document}